# Classifier Suites for Insider Threat Detection

David A. Noever, Sr. Technical Fellow,
PeopleTec, Inc. 4901-D Corporate Drive, Huntsville, AL 35805 USA

**Abstract**

Better methods to detect insider threats need new anticipatory analytics to capture risky behavior prior to losing data. In search of the best overall classifier, this work empirically scores 88 machine learning algorithms in 16 major families. We extract risk features from the large CERT dataset, which blends real network behavior with individual threat narratives. We discover the predictive importance of measuring employee sentiment. Among major classifier families tested on CERT, the random forest algorithms offer the best choice, with different implementations scoring over 98% accurate. In contrast to more obscure or black-box alternatives, random forests are ensembles of many decision trees and thus offer a deep but human-readable set of detection rules (>2000 rules). We address performance rankings by penalizing long execution times against higher median accuracies using cross-fold validation. We address the relative rarity of threats as a case of low signal-to-noise (< 0.02% malicious to benign activities), and then train on both under-sampled and over-sampled data which is statistically balanced to identify nefarious actors.

## Introduction

Industry surveys asking "What do you consider the greatest security threat to your organization?" have shown that 77% of respondents topped their list with their own employee's actions (either careless [60%] or disgruntled [17%] ones) (Dark Reading, 2014). However detection of trusted individuals with authorized credentials doing malicious things has been classified as one of the hardest cybersecurity problems (Hunker, et al, 2010) (and second only to nation-state cyberattack attributions, see Noever, et al., 2016). Therefore one motivation for this work is a growing realization that insider threat detection is neither looking at the right risk indicators nor applying state-of-the-art algorithms with common scoring metrics (Greitzer, et al. 2010).

We apply novel feature engineering, automatically score each factor's relevance and then test the classification success using a large machine learning suite of 88 algorithms in 16 major classifier families. A rough guide to which classifiers best apply in shown in Figure 1 based on problem features, sample number and whether an explainable reasoning requirement is needed. The last step follows similar comprehensive surveys by Caruna, et al. (2006) and Fernández-Delgado, et al. (2014); the latter particularly compared many learning algorithms (179) across diverse datasets (121). We specialize these algorithms to insider threat detection and identify particular weak and strong performers for both classifier families and their key predictive features. The previous work by Caruna, et al. (2006) tested 8 model families (including 2000 variations of tuning parameters) against 7 binary classification problems (mainly medical) and found boosted decision trees as the best overall classifier. They also found good performance overall for neural nets, support vector machines, and bagged trees, depending on the scoring metric. Fernández-Delgado, et al. (2014) expanded this survey of best algorithms to 179 classifiers in 17 families against 121 datasets, the entire UCI machine learning data repository. They also found random forests as the most successful one overall. Finally massive online competitions such as Kaggle have attracted thousands of data scientists to apply their favorite algorithms and independently score them against industry-relevant problems. Three key anecdotal results from Kaggle (Goldbloom, 2015) emphasize the power of combining multiple algorithms, good feature engineering, and the general success of ensembles of decision trees (particularly random forests and extreme gradient boosting). In contrast to black-box alternatives with obscure interpretative rules, random forests are ensembles of many decision trees and thus offer a deep but human-readable set of detection rules. Such explainable artificial intelligence is recognized as a growing need and the subject of intense investigations particularly in high-stakes decision realms like medicine, defense, and human resources.

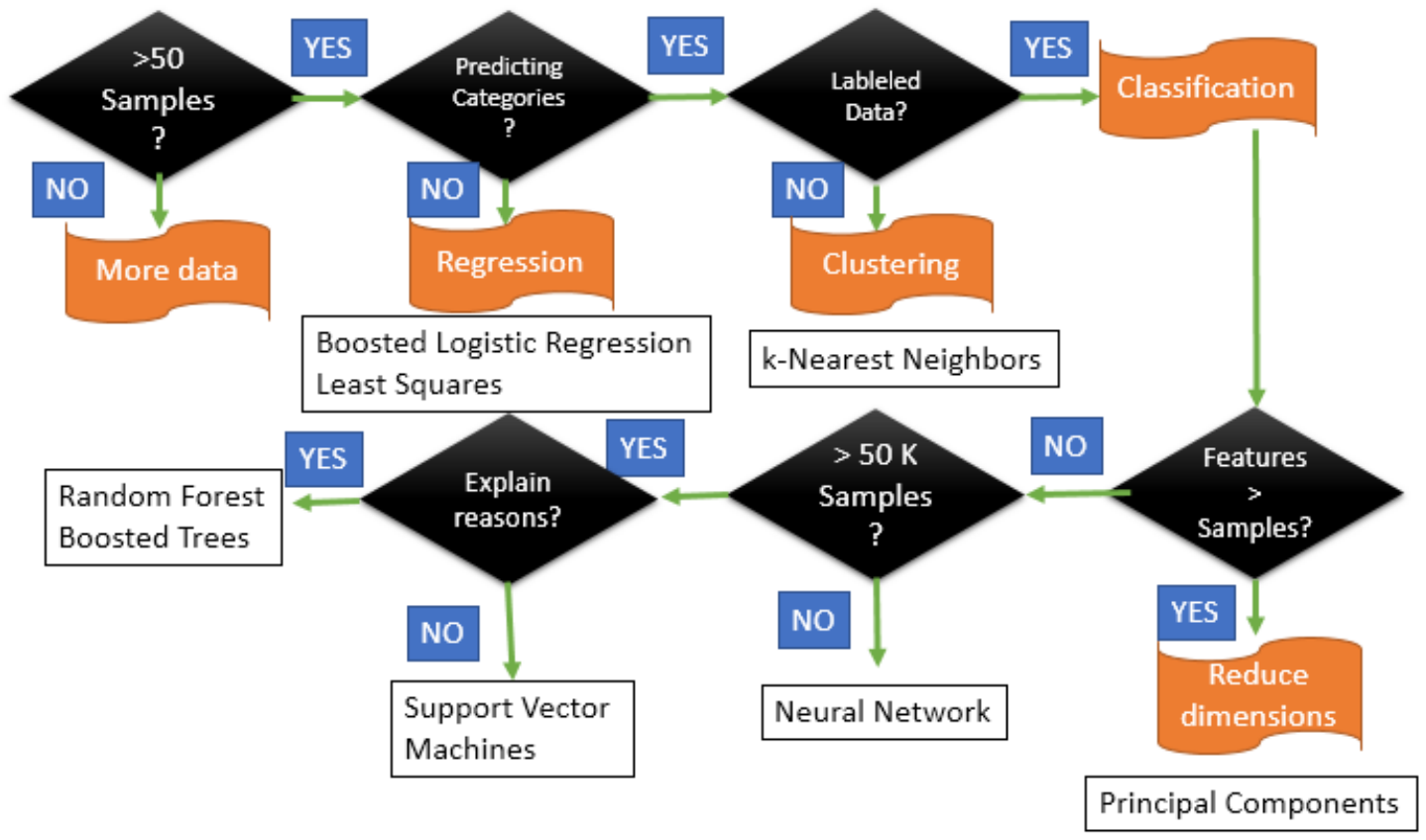

*Figure 1 Rough Guide to Choosing the Best Machine Learning Families*

**Methods**

The Carnegie Mellon University Software Engineering Institute's CERT Program's Insider Threat database provides the largest public repository of red team scenarios. The simulation derives veracity from blending real-world insider threat case studies with actual benign users sampled anonymously from a defense corporation. The terabyte-sized data set describes 18 months of observing traffic from web, email, logon, file and device access (Table 1) in a single engineering company (dtaa.com). The mock company employs between 1000 and 4000 people who each perform an average 1000 logged activities per day. We address the top three of five total threat scenarios (leaker, thief, saboteur excluding data hoarders and laid-off Dropbox cases). The data set comprises 209 million raw data points and 32 million time-stamped actions. The social connections between paired (to/from) users offer a half-million unique connections among just the top 3% of email exchanges. A notable feature of the problem is its very low signal-to-noise whether measured in total malicious users, daily tallies, or total usage: 0.2% of users (99.8% users benign); 0.08% of user-days (99.92% daily usage benign); 0.02% of user-actions (99.98% total usage benign).

*Table 1 CERT Dataset Activity, Features and Scale*

| Activity | Number | Feature Count | Features Description |
|---|---|---|---|
| **Website** | 28,434,423 | 6 | id,date,user,pc,url,content |
| **Email** | 2,629,980 | 11 | id,date,user,pc,to,cc,bcc,from,size,attachments,content |
| **Logon** | 854,860 | 5 | id,date,user,pc,activity [logon/off] |
| **Files** | 445,581 | 6 | id,date,user,pc,filename,content |
| **Device** | 405,380 | 5 | id,date,user,pc,activity [connect/disconnect] |

In additions to activity logs, we augment the CERT dataset with other transformation metrics and outside supplemental sources. An example data transformation is to pair employee-supervisor roles as a factor. Examples of supplemental data include employee salary by job title, gender by first name, web categories by blacklists. To fill in features derived from the original CERT usage logs, Table 2 shows our novel data augmenting sources.

*Table 2 Data Augmentation Strategy and Choices for Predictive Feature Extraction*

| Feature | Benefit | Source |
|---|---|---|
| **Employee-Supervisor Network** | Role Evolves over Time | Calculated as Network Pairs |
| **Web Domain Type** | Categorical Assignments | URLBlacklist.com |
| **Gender by First Name** | Behavioral/ Contextual | Social Security Historical |
| **Salary by Job Title** | National Averages | Indeed.com |
| **Sentiment Analysis** | Email/File/Web Gradations | AFINN-111 |
| **Date Manipulations** | Day of Week, Hour of Day | Calculated |
| **Statistical Features** | Change of life patterns | Calculated |
| **Outlier Features** | Change of life patterns | Calculated |
| **Log file Parsers** | Summations, Event Memory | Calculated |
| **Human Resource Analytics** | Behavioral/ Contextual | Calculated |

**Dataset Preprocessing and Feature Extraction**

This analysis centered on one CERT Insider Threat scenario (v.4.2) specifically because it represents the only 'dense needle' dataset. Among the 6 total threat scenarios otherwise featuring low threat counts or 'needle in a haystack' problems, scenario v.4.2 includes approximately 7% of the 1000 employees as nefarious for at least 1 of 18 month of logging. The alternative CERT scenarios follow similar threat narratives but provide just single cases for each insider class (thief, saboteur and leaker). However even this denser 4.2 scenario underscores the unbalanced nature of insider threat datasets generally, even with its 93% benign activity rates (see Lindauer, et al. 2014). Therefore we balanced the classification problem with statistical under-sampling of benign behavior and over-sampling of nefarious behavior. Because CERT reports critical job events such as employee departures as monthly changes, we extracted features from web, email, file and device logs across a similar monthly window. We calculate monthly risk factors to summarize indicators for insider threat detection. Notably risky examples included incident counts for any outliers working outside normal hours and weekends.

*Table 3 Predictive Risk Indicators Extracted as CERT Features*

| Risk Indicator | Description |
|---|---|
| **risk_leak** | Leak risk calculated as event counts for email or web logs to file sharing domains |
| **dow_leak** | Day of week (weekend) count for leak risks |
| **hr_leak** | Off-hour (outside 8 AM to 5 PM) count for leak risks |
| **risk_thief** | Theft risk calculated as event counts for email or web logs to job search or competitor domains |
| **dow_thief** | Day of week (weekend) count for theft risks |
| **hr_thief** | Off-hour (outside 8 AM to 5 PM) count for theft risks |
| **risk_sabotage** | Sabotage risk calculated as event counts for web logs to malware or keylogger domains |
| **dow_sabotage** | Day of week (weekend) count for sabotage risks |
| **hr_sabotage** | Off-hour (outside 8 AM to 5 PM) count for sabotage risks |
| **device_freq** | Monthly device frequency (e.g. USB usage) |
| **file_freq** | Monthly file frequency (e.g. Word, PDF) |
| **email_sentiment** | Calculated monthly sentiment (polarity as positive or negative) in email text |
| **email_compete** | Calculated monthly count of emails to competitors |
| **web_sentiment** | Calculated monthly sentiment (polarity as positive or negative) in visited website text |
| **executables** | Calculated monthly file access to executables |
| **unauthorized_log** | Unauthorized device usage after employee departure |
| **gender** | Inferred employee gender from first name |

As shown in Table 3, we tabulated these threat features for each employee, along with their monthly class as: benign, departed, leaker, thief or saboteur. A single employee can reside in more than one class, since all begin as benign. For a baseline 1000 employees over 18 months, the maximum sample therefore reduces to 18,000 training cases. Other than benign employees (80%) and former (departed) employees (13%), the nefarious cases are distributed by employee count as thieves (3%), leakers (3%) and saboteurs (1%). From the larger universe of features that are domain-specific such as activities outside normal work hours, the highlighted features were visualized as statistical anomalies from pre-threat behavior. As shown in Figure 2, a box-whisker chart shows some of the more interesting class attributes compared to the company as a whole. For the leaker and saboteur classes, other risky indicators combined with weekend or after-hours access can guide a potentially successful class partition. Distinguishing the thief class however proves more challenging, given that as part of their daily routines, the benign users also typically email competitors, visit job sites and access networks outside normal hours.

We analyze sentiment factors in email, website content and accessed file text in a multistep process. We account for single word stems by removing numbers, stripping punctuation and lower-casing the text. For English sentiment, we employ the AFINN-111 (Nielsen, 2011) list of single words or phrases (2477) labelled on an integer (1-5) scale for positive or negative polarity. Examples of strong negative sentiment include most curse words, racial or class epithets and grievances (e.g. “tortures”). Examples of strong positive sentiment include broad approvals (e.g. “breakthrough” or “outstanding”).

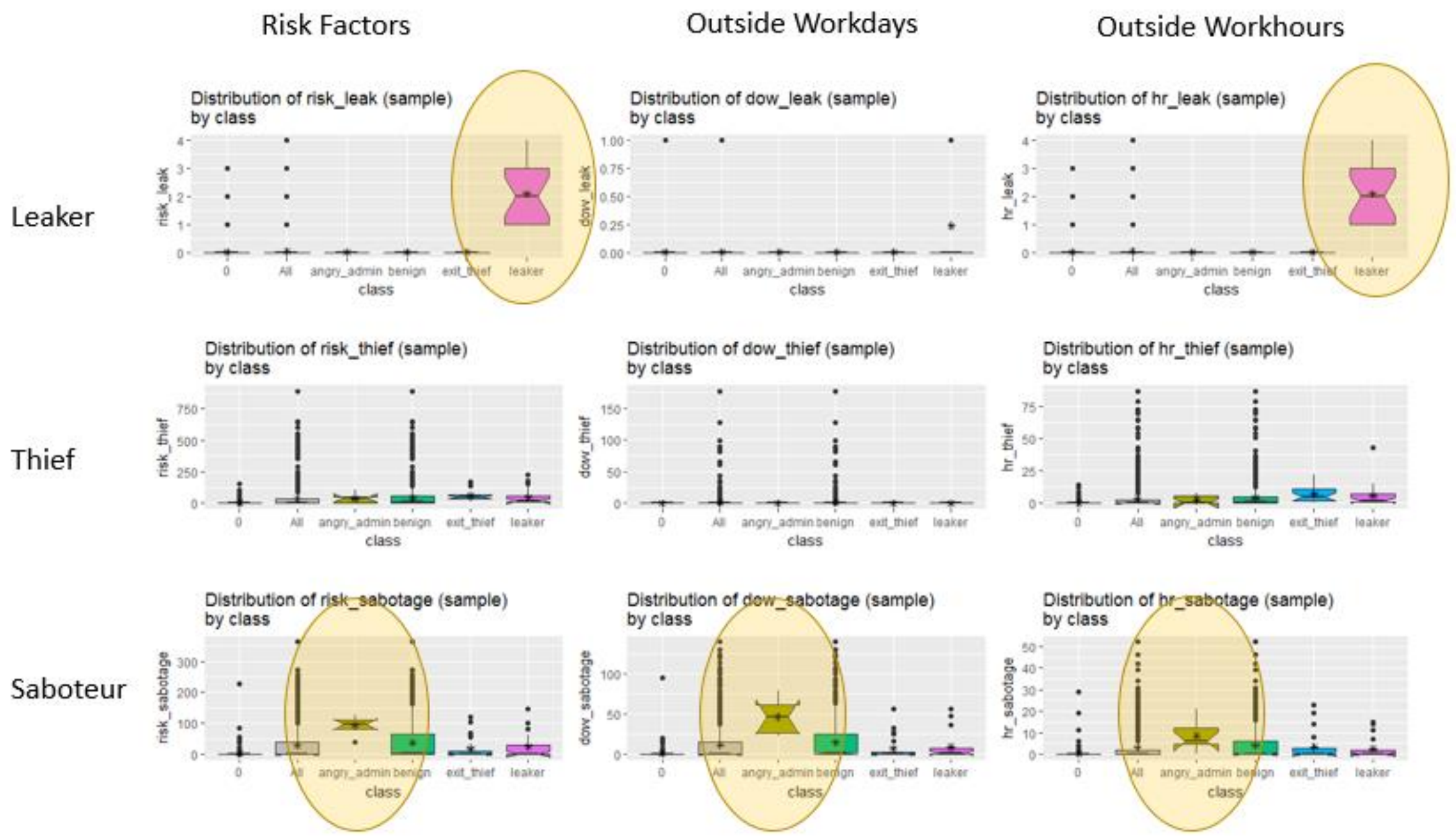

*Figure 2 Box-Whisker Chart Highlights the Risk Distribution across Behavioral Classes*

**Description of Feature Selection**

We applied automated feature selection to rank order each factor’s importance to identify each class. The approach assigns a top-down search based on Boruta as in Miron, et al. (2010). The approach systematically removes each factor and finds the loss of information in the final prediction. This algorithm confirms that each factor is statistically predictive and that none can be rejected as unimportant. Interestingly in the CERT data, inferred gender is relevant but the least predictive among the 17 features. Consistent with the previous findings shown in Figure 2, the least important factors in Figure 3 are also not the commonly logged activity frequencies such as access to executables, files and devices. This suggests that looking for employee-specific changes in these frequencies may not classify CERT abnormalities following the simpler workplace rules like banning USB devices or restricting executables.

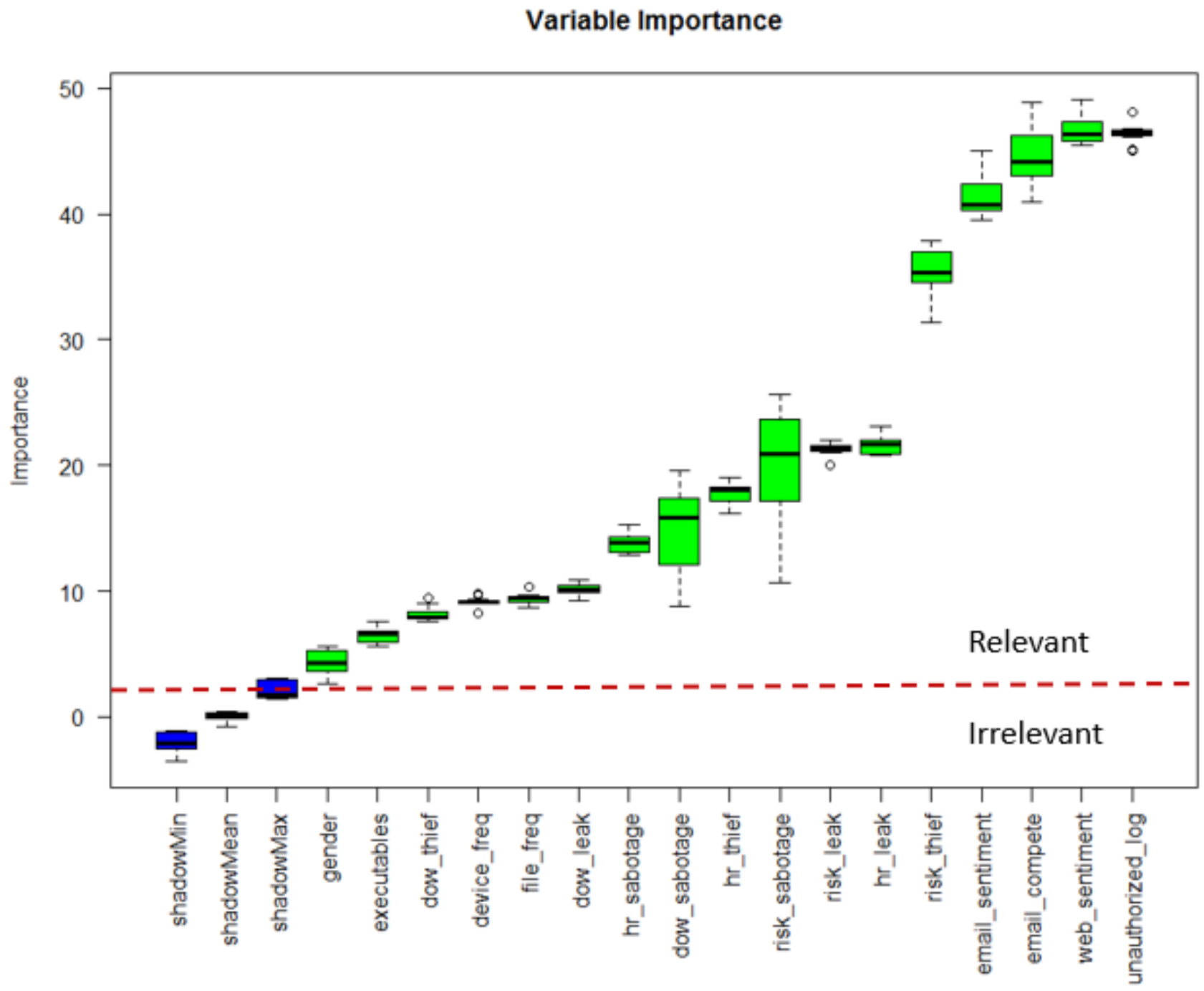

*Figure 3 Rank Order of Predictive Information Embedded by Each Risk Feature*

The broad feature contribution of employee sentiment in email and visited websites suggests a data pre-processing step might outperform single-factor business rules for designing or deploying new risk indicators. It is worth noting that CERT generously provides text for all employee visited websites which presents considerable practical challenges to archive and track continuously for even moderately sized enterprises. CERT may have provided this website text to guide researchers to create their own domain blacklists, which now routinely ship as updated, community supplements with all modern security software. As described in Table 2, we used the 4 million community-categorized domains from URLBlacklist.com. Such website domain and categorization lists help to identify risky sites for employee interactions with competitors, webmail, file sharing, malware, and job searches. It is worth noting however that most benign employee actions also routinely access these domains (particularly webmail, competitors and job searches) and thus caution is needed if expecting them to be predictive (as enforced by company policies or standalone detection rules).

**Description of Classifier Suite**

Following data pre-processing and feature extraction, we undertook a survey of the machine learning landscape for handling a five-class partition (benign, departed, thief, leaker and saboteur). The monthly data was anonymized by employee id, such that each instance represents a one-shot prediction with no knowledge of employee id or their previous cumulative behavior. For the unbalanced 18,000 training instances, the over-sampled set magnified the nefarious actors to provide equal distributions (78,446) for each class without distorting the statistical properties of input features. Similarly the under-sampled set reduced the benign actors to similarly provide only 10 cases each (50 total) but with representative input properties. The motivation for this statistical treatment stems partly from unbalanced training where algorithms converge prematurely to predict all employees act benignly (low true negatives). In practice, however, most currently deployed threat algorithms compound this bias by over-tuning their risk indicators and thus yield very high false positives and unjustifiably flag benign employees. CERT (2017) concludes "many insider threat detection tools produce so many false positives that the tools are unusable." A common user complaint about security software thus centers on the need either to chase down employees who pose no threat or to ignore the indicators altogether in favor of administrative intuition and judgement.

We implemented the large-scale classifier suite using the R algorithm libraries conveniently packaged in a wrapper library called caret. This machine learning wrapper described by (Kuhn, 2008) provides a common calling function, train, which simplifies such large scale algorithm surveys. Currently caret supports 179 algorithms as classifiers. For insider threat detection, our final machine learning suite included 88 algorithms in 16 major classifier families for performing regression (linear or logistic multinomial, adaptive spline, partial least squares, polynomial); Bayesian; neural networks including deep and shallow layers; support vector machines; rule-based decision trees including ensembles such as boosting, bagging, stacking and random forests. We ranked the algorithms on two often competing axes: median predictive accuracy and relative execution time. Since execution time is machine-dependent, relative ranking and log rescaling between 0-1 proved the most useful time metric. We validated performance by partitioning training and testing data, along with repeated cross-fold validation in both the under-sampled and over-sampled cases. We attempted no parameter tuning in favor of accepting model defaults. Models were excluded if they took longer than a few hours or failed to converge, a compromise which accounted for about half the attrition from the original list of 179 algorithms.

**Results**

Figures 4 and 5 summarize the classifiers ranked from best to worst for the over-sampled and under-sampled training sets respectively. Among major classifier families, the best raw scores derive from the random forest algorithms, with different implementations scoring over 98% accurate in the over-sampled dataset with 78,446 cases and 94% accurate in the under-sampled dataset with 50 cases. As a general out-performing algorithm, the success of random forests mirrors the previous findings in large-scale surveys and international data competitions like Kaggle.

To emphasize the potential tradeoff between execution speed and accuracy, we rescaled the relative time using log(time)/max(log(time) and plotted the Accuracy vs. Time scatter to visualize the faster and more accurate algorithms in Figure 6. with classifier family color-coded. For instance the upper left quadrant represents one view of the faster, better family of choices and the lower right corresponds to slower, poorer choices. In practice, the classifier choice may also depend on data size, noise and need for continuous vs. batch training.

For the better performing random forest family, Figure 7 shows an error- or confusion-matrix for each insider threat cases where actual vs. predicted threats fall into the four quadrants (e.g. true/ false positive/negative cases). All the classes show false assignments (red quadrant) in less than 1% of the cases. Consistent with the original box-whisker charts in Figure 2, predicting the thief's behavior presents the biggest challenge as the thief's activities resemble those of benign employees but even in this more overlapping class, the best algorithms can still classify them with 1.1% false assignment rates.

| Method | Accuracy | Kappa | Time | Model | Method | Accuracy | Kappa | Time | Model |
|---|---|---|---|---|---|---|---|---|---|
| ranger | 0.9835 | 0.9661 | 210.56 | Random Forest | svmLinearWe | 0.9111 | 0.8147 | 131.92 | Linear Support Vector Machines with Class Weights |
| extraTrees | 0.9831 | 0.9651 | 200.16 | Random Forest by Randomization | glmnet | 0.9087 | 0.8097 | 43.5 | glmnet |
| knn | 0.9814 | 0.9618 | 4.44 | k-Nearest Neighbors | glm | 0.9077 | 0.8081 | 2.06 | Generalized Linear Model |
| loclda | 0.9806 | 0.9601 | 114.64 | Localized Linear Discriminant Analysis | multinom | 0.9027 | 0.7976 | 6.49 | Penalized Multinomial Regression |
| RRFglobal | 0.975 | 0.9483 | 475.19 | Regularized Random Forest | amdai | 0.8779 | 0.7409 | 1.07 | Adaptive Mixture Discriminant Analysis |
| treebag | 0.9727 | 0.9438 | 18.66 | Bagged CART | rda | 0.8777 | 0.7406 | 10.85 | Regularized Discriminant Analysis |
| LMT | 0.9683 | 0.9347 | 49.61 | Logistic Model Trees | lvq | 0.8544 | 0.6895 | 108.11 | Learning Vector Quantization |
| gbm | 0.9675 | 0.933 | 13.38 | Stochastic Gradient Boosting | rpartCost | 0.8309 | 0.6652 | 2.28 | Cost-Sensitive CART |
| bagEarthGCV | 0.9656 | 0.9289 | 358.49 | Bagged MARS using gCV Pruning | kknn | 0.8185 | 0.6171 | 26.6 | k-Nearest Neighbors |
| JRip | 0.9652 | 0.9282 | 178.05 | Rule-Based Classifier | avNNet | 0.8066 | 0.5762 | 50.69 | Model Averaged Neural Network |
| bagEarth | 0.9646 | 0.927 | 651.31 | Bagged MARS | qda | 0.8038 | 0.5672 | 1.08 | Quadratic Discriminant Analysis |
| J48 | 0.9615 | 0.9206 | 21.2 | C4.5-like Trees | rpart2 | 0.8014 | 0.5626 | 2.03 | CART |
| PART | 0.9571 | 0.9117 | 24.39 | Rule-Based Classifier | naive_bayes | 0.7993 | 0.5581 | 1.15 | Naive Bayes |
| gcvEarth | 0.9516 | 0.9001 | 11.05 | Multivariate Adaptive Regression Splines | nb | 0.7989 | 0.5573 | 8.41 | Naive Bayes |
| ada | 0.9504 | 0.8969 | 115.4 | Boosted Classification Trees | rpart | 0.7814 | 0.5133 | 2.31 | CART |
| monmlp | 0.95 | 0.897 | 454.47 | Monotone Multi-Layer Perceptron Neural Network | rpartScore | 0.7814 | 0.5133 | 120.84 | CART or Ordinal Responses |
| bstTree | 0.9455 | 0.8868 | 55.95 | Boosted Tree | sdwd | 0.7799 | 0.5083 | 51.22 | Sparse Distance Weighted Discrimination |
| rpart1SE | 0.9361 | 0.8679 | 3.13 | CART | nnet | 0.7266 | 0.3785 | 12.87 | Neural Network |
| LogitBoost | 0.9341 | 0.8641 | 8.69 | Boosted Logistic Regression | bstSm | 0.7027 | 0.3141 | 786.03 | Boosted Smoothing Spline |
| svmLinear | 0.9111 | 0.8147 | 14.72 | Support Vector Machines with Linear Kernel | rocc | 0.6971 | 0.3009 | 2.85 | ROC-Based Classifier |
| svmLinear2 | 0.9111 | 0.8147 | 51.35 | Support Vector Machines with Linear Kernel | pls | 0.6765 | 0.3033 | 1.22 | Partial Least Squares |
| | | | | | dnn | 0.5933 | 0.0255 | 173.97 | Stacked AutoEncoder Deep Neural Network |

*Figure 4 Over-sampled Results by Model Accuracy and Time*

As a proxy simple rule-based detectors, the closest listed comparison is the recursive partition (or RPART) algorithm which achieves 78% (over-sampled) and 38% (under-sampled) accuracy rates. This industry-standard approach is a multi-branch "if/then" decision tree. Its deployment is partly motivated by its ease of interpretation and user modifications particularly when implemented as single-depth trees (no branching for other factors). However for complex, multi-featured datasets such as insider threats, the tendency for simple decision trees to overfit training data and predict poorly leads to high false-positive rates. Simple rule-based detectors lag even further when scoring the typically unbalanced training data that threat prediction demands post-deployment. A best case of rules-based classifiers achieving 20% error rates while evaluating a 1000-person company would present a practical challenge to its cybersecurity team.

| Method | Accuracy | Kappa | Time | Model |
|---|---|---|---|---|
| LogitBoost | 0.96 | 0.95 | 0.72 | Boosted Logistic Regression |
| extraTrees | 0.94 | 0.925 | 1.62 | Random Forest by Randomization |
| LMT | 0.92 | 0.9 | 1.28 | Logistic Model Trees |
| ranger | 0.9 | 0.875 | 0.69 | Random Forest |
| RRFglobal | 0.9 | 0.875 | 0.83 | Regularized Random Forest |
| PART | 0.9 | 0.875 | 0.86 | Rule-Based Classifier |
| J48 | 0.9 | 0.875 | 1.78 | C4.5-like Trees |
| treebag | 0.88 | 0.85 | 0.78 | Bagged CART |
| kknn | 0.82 | 0.775 | 0.63 | k-Nearest Neighbors |
| glmnet | 0.82 | 0.775 | 0.77 | glmnet |
| JRip | 0.78 | 0.725 | 1.8 | Rule-Based Classifier |
| monmlp | 0.78 | 0.725 | 1.8 | Monotone Multi-Layer Perceptron Neural Network |
| svmLinear2 | 0.76 | 0.7 | 0.53 | Support Vector Machines with Linear Kernel |
| svmLinear | 0.76 | 0.7 | 0.61 | Support Vector Machines with Linear Kernel |
| avNNet | 0.7 | 0.625 | 0.94 | Model Averaged Neural Network |
| multinom | 0.66 | 0.575 | 0.57 | Penalized Multinomial Regression |
| nnet | 0.64 | 0.55 | 0.61 | Neural Network |
| rda | 0.6 | 0.5 | 0.66 | Regularized Discriminant Analysis |
| pls | 0.58 | 0.475 | 0.55 | Partial Least Squares |
| naive_bayes | 0.48 | 0.35 | 0.64 | Naive Bayes |
| lvq | 0.42 | 0.275 | 0.65 | Learning Vector Quantization |
| rpartScore | 0.4 | 0.25 | 1.03 | CART or Ordinal Responses |
| rpart1SE | 0.38 | 0.225 | 0.53 | CART |
| knn | 0.38 | 0.225 | 0.54 | k-Nearest Neighbors |
| rpart2 | 0.38 | 0.225 | 0.61 | CART |
| rpart | 0.38 | 0.225 | 0.66 | CART |
| bstTree | 0.36 | 0.2 | 2.91 | Boosted Tree |
| nb | 0.34 | 0.175 | 0.77 | Naive Bayes |
| dnn | 0.32 | 0.15 | 2.23 | Stacked AutoEncoder Deep Neural Network |
| rocc | 0.3 | 0.125 | 0.67 | ROC-Based Classifier |

*Figure 5 Under-sampled Results by Model Accuracy and Time*

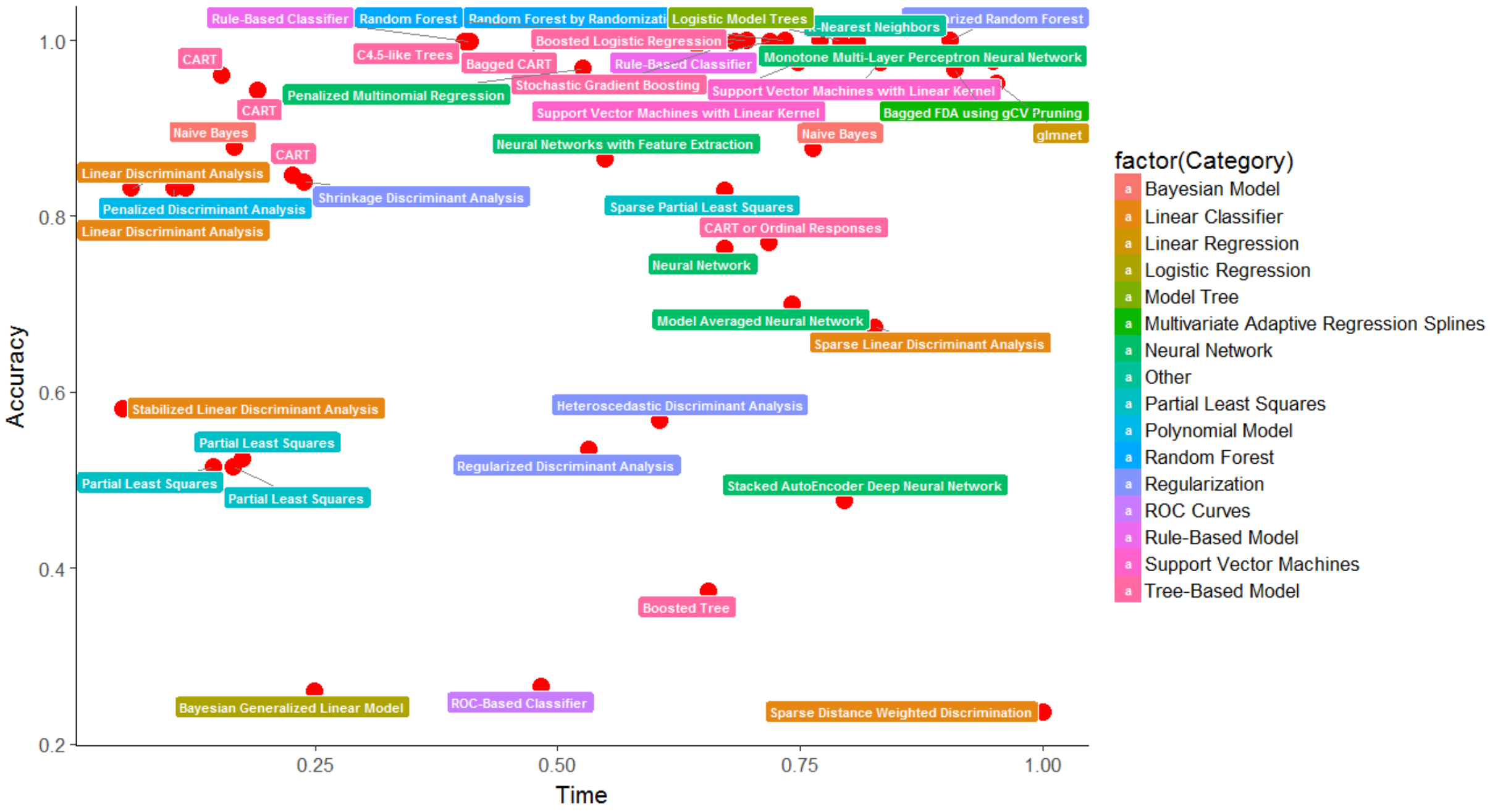

*Figure 6 Algorithm accuracy as a function of execution times*

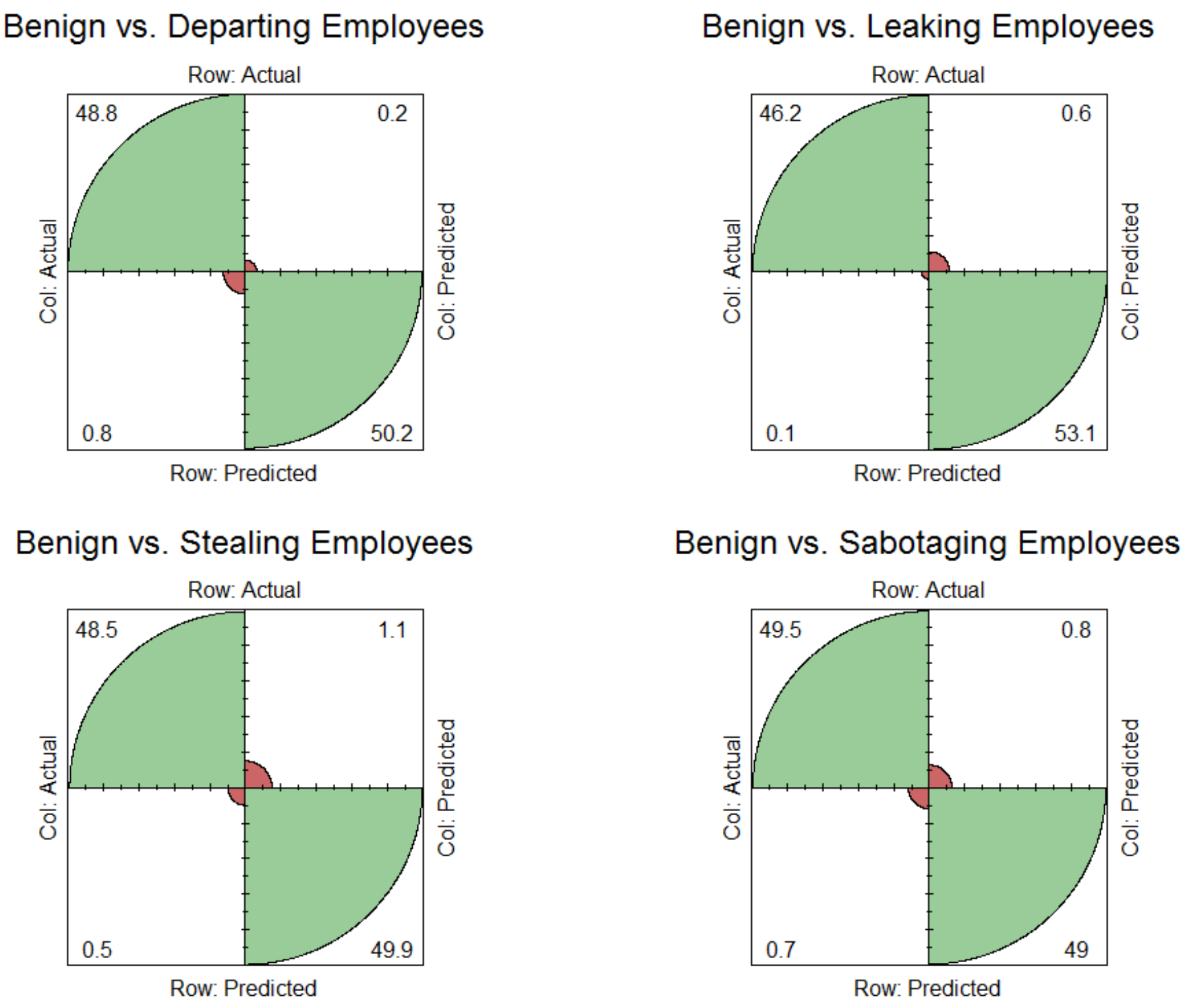

*Figure 7 Error or confusion matrices for insider threat classes.*

**Discussion and Conclusions**

A surprising result is the broad success found here applying random forests to detect insider threats. Other authors have employed random forests for detecting anomalous behavior but only for particular subsets of online activity in games like World of Warcraft, social media or modeled datasets from other realms (Brdiczka, et al, 2012; Gavai, et al. 2015; Emmott, et al. 2015). To our knowledge, none has applied systematic algorithmic scoring to the CERT dataset, or included rankings of new features by their predictive importance. A number of commercial tools deploy rule-based, online monitoring but do not provide guidance on their ability to generalize or learn, or their objective scoring against some ground truth.

**A data-driven challenge to single-rule detectors:** The present approach relies on collecting both accepted and novel metrics or predictive risk indicators (PRIs) but differs in its independence from peer-group observation or simple rule-based detection. One way to compare these machine learning results to existing rule-based security systems is to take an industry (Deloitte) list of top 25 predictive risk indicators (PRI) and score them against the CERT dataset. Many of the PRIs in Table 4 highlight compliance violations, which in many cases can be eluded by a cautious insider or routinely flouted by unknowing but benign users. For illustration, we rank order the 18,000 user-months by each PRI and assume that even an ambitious IT staff can track only the top 100 cases over time. We focus on 6 of the 25 PRIs that involve CERT-available network logs, including file, malware, and device activity along with emailing competitors and web sentiment (as a proxy for social media anomalies). All 6 network PRIs lead to extremely high false positives (84-100%) and low true negatives (0-23%). One reason for this failure may derive from CERT blending real network usage with threat narratives and the low signal-to-noise in real-world detection. Another reason may follow from individual behavioral changes dominating the deviations from the enterprise mean. A final reason may stem from not including the critical threat triggering event, usually a work termination or dispute which dramatically alters the importance of a given factor (e.g. collecting many files after a grievance). The IT challenge however remains when building risk dashboards, for instance, if they show the daily top 100 risky employees and then task the IT staff to discover the root causes. What may look like poor employee behavior may arise more clearly as just poor metrics and underperforming algorithms.

The often cited answer to these challenges is to encourage employee self-patrols, a policy of encouraging internal amateur profiling. This approach requires a lot of sophistication and may arise more as a recognition of the data-mining challenges than a reliable threat detector. For example false positives arise routinely from employees who are sick, change job roles or cope with unknown outside challenges (medical, etc.). Costly false negatives can also arise from organizations that advertise detection criteria to the often clever thief in the hopes that co-workers can sift through the subtle or hidden changes in co-workers' abnormal weekend events, after-hour access, and the explosion of compliant but malicious copying strategies (phones, cameras, and multimedia).

This low success rate from applying business rules follows what FBI guidance has previously found, namely that only 5% of its 65 espionage cases were correctly flagged as potential bad actors and their first predictive models did worse than random (Ready, 2013). A key reason cited was unbalanced datasets with an excess of noisy (but benign) events and the lack of true positives to train detectors on rare or "Black Swan" events. A further qualification was that outliers must arise from individual patterns of activity and not deviations from average enterprise monitoring. This focus on individual deviations is akin to distinguishing a normal anomaly from an abnormal one. Ready suggests five broad insider threat approaches: 1) assume employees don't behave like outside hackers; 2) assume the threat is not a technical issue alone or a particularly good candidate for intrusion and antivirus monitors; 3) limit the data overload inherent in the normal, benign behavior; 4) educate the workforce to focus on deterrence and crowd-sourcing detection; and 5) use behavioral analytics to prioritize risky changes in an individual's pattern of life. The latter issue was examined here to assess the effectiveness of reducing false positives and predicting data theft or leakage and IT sabotage.

*Table 4 Top 25 Industry Risk Indicators and Accuracy as Single Rules against CERT Data*

| Predictive Risk Indicator | False Accusations | Eludes Detection | Metric |
|---|---|---|---|
| **Collection of large quantities of files** | 100% | 100% | File Frequency |
| **Antivirus/malware alerts** | 100% | 100% | Risk Sabotage |
| **Excessively large downloads** | | | N/A |
| **Access request denials** | 84% | 77% | Unauthorized Logons |
| **Large outbound e-mail traffic volume** | | | N/A |
| **Emails with attachments sent to suspicious recipients** | 100% | 100% | Email Competitors |
| **Transmittal device (e.g. printers, copiers, fax machine) anomalies** | | | N/A |
| **Removable media alerts and anomalies** | 99% | 98% | Device Frequency |
| **Access levels** | | | N/A |
| **Security clearance** | | | N/A |
| **Privilege user rights** | | | N/A |
| **Physical access request denials** | | | N/A |
| **Physical access anomalies** | | | N/A |
| **Audit remediation progress** | | | N/A |
| **Non-compliance with training requirements** | | | N/A |
| **Organizational policy violation (e.g. data classification policy, avoiding an e-QIP)** | | | N/A |
| **Expense violations** | | | N/A |
| **Time entry violations** | | | N/A |
| **Declining performance ratings** | | | N/A |
| **Notice of resignation or termination** | | | N/A |
| **Reprimand or warning** | | | N/A |
| **Social media anomalies** | 100% | 100% | Web Sentiment |
| **Financial stressors** | | | N/A |
| **Criminal and Civil History Background Checks** | | | N/A |
| **Foreign contacts/travel** | | | N/A |

**The potential for better natural language processors:** Given the apparent predictive power of sentiment analysis, it is worth mentioning that in the interest of processing speed, we scored only the simplest polarities. Many alternative dictionaries can be compared, along with more contextual multi-word (or n-grams), but the initial choice here was to provide the cumulative monthly sentiment total for each employee. A fuller survey of machine learning methods for natural language scoring would add many more dimensions to ranking these classifiers but one might incorporate the best methods in a two-stage classifier that first extracts sentiment features, then predicts the employee classification. The challenge to that approach however is that lack of a pre-labelled training and testing set for sentiment scores that might guide the a best selection criterion.

**The potential for unsupervised learning methods:** A commonly cited hindrance to building better insider detectors is the lack of more training datasets that are pre-labeled with threat classes (see Emmott, et al, 2013). One result found here related to this challenge is that a few good algorithms can learn from unlabeled threat data. For such unsupervised learning, network traffic for individuals are clustered by activity types without manual labels or pre-built threat narratives. The advantages of unsupervised learning derive from automated clustering in high-dimensional problem spaces. Unsupervised learning also can remove the cumbersome and often inaccurate data generating steps for labelling and updating to evolving threats. In this regard, two algorithm families (k-Nearest Neighbors and Stacked Auto-encoder Deep Neural Networks) deserve particular mention in our classifier rankings, since they can potentially operate unsupervised without knowing an employee's threat rating. While deep neural nets offer great promise in other application fields for classifying large image or audio datasets, one surprising result is they generally fail against the CERT data. Deep neural nets perform poorly in the bottom 10% of algorithms, perhaps because of the unscaled input ranging approximately from 0-1000. Tuor, et al. (2017) recently reported promising 95.53% anomaly detection using many deep nets trained individually for each user in CERT v.6.2. For nearest neighbors, however, its clustering performs well for this multi-class problem (scoring in the top 10% both in speed and accuracy). It deserves further investigation as an unsupervised learning candidate for accommodating new, previously unseen threats and finding probable threat types. Future work will apply k-nearest neighbors for labeling. A next step for exploiting the success of a fully-trained random forest can focus on reducing these multi-featured decision trees to a high-speed "if/then" filtering algorithm and SQL-like queries to run against real-time online activities.

**Acknowledgements**
The authors would like to thank the PeopleTec Technical Fellows program for encouragement and project assistance.